\documentclass[a4paper, 10pt, conference]{ieeeconf}      
\usepackage{FG2026}

\FGfinalcopy 

\IEEEoverridecommandlockouts        
\usepackage[whole]{bxcjkjatype}
\usepackage{graphicx}
\usepackage{amsmath}
\usepackage{amssymb}
\usepackage{booktabs}
\usepackage{bm}
\usepackage{url}
\usepackage[pagebackref,breaklinks,colorlinks]{hyperref}

\usepackage{xcolor}      
\usepackage[normalem]{ulem}   

\newcommand{\del}[1]{\bgroup\markoverwith{\textcolor{red}{\rule[0.5ex]{2pt}{0.4pt}}}\ULon{#1}}

\usepackage[capitalize]{cleveref}
\crefname{section}{Sec.}{Secs.}
\Crefname{section}{Section}{Sections}
\Crefname{table}{Table}{Tables}
\crefname{table}{Tab.}{Tabs.}

\input{mlpreamble}

\newcommand{\cmas}{cMAS\xspace}

\def\FGPaperID{7} 

\usepackage{fancyhdr}

\title{\LARGE \bf
Unsupervised 3D Human Pose Estimation \\via Conditional Multi-view Ancestral Sampling
}

\author{\parbox{16cm}{\centering
   {\large Ryohei Goto \quad Takuya Fujihashi \quad Shunsuke Saruwatari \quad Fumio Okura}\\
   {\normalsize
   The University of Osaka\\}}
}

\begin{document}

\ifFGfinal
\thispagestyle{empty}
\pagestyle{empty}
\else
\pagestyle{plain}
\fi
\maketitle
\thispagestyle{fancy}

\begin{figure*}[t]
        \centering
        \includegraphics[width=\linewidth]{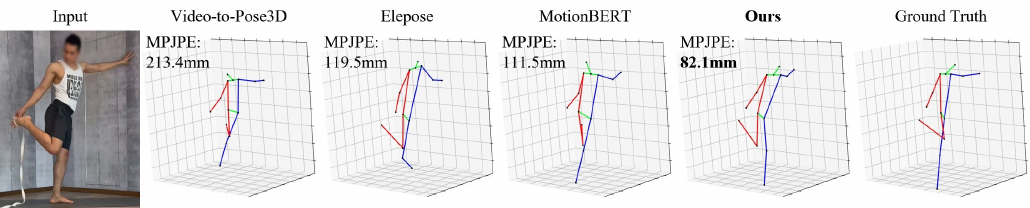}\vspace{-2mm}
        \caption{Compared to the baselines including state-of-the-art supervised (Video-to-Pose3D~\cite{video-to-pose3d} and MotionBERT~\cite{MotionBERT}) and unsupervised (ElePose~\cite{ElePose}) methods, our \cmas-based method achieves accurate 3D pose estimation even for extreme poses (\eg, yoga), leveraging a rich 2D diffusion prior and its flexibility to incorporate additional heuristic priors, such as anatomical constraints.
        \vspace{-3mm}
        } \label{fig:teaser}
\end{figure*}

\begin{abstract}
We propose a method of estimating a 3D human pose from a single view without 3D supervision. The key to our method is to leverage the 2D diffusion priors of motion diffusion models (MDMs) pre-trained on large 2D human pose datasets. Specifically, we extend multi-view ancestral sampling of diffusion models to the task of 2D-3D lifting of human pose. To this end, we newly propose a \emph{conditional multi-view ancestral sampling (\cmas)} that optimizes the 3D pose such that its multi-view projections follow the manifold in 2D MDM noise space, while conditioning the 3D pose to match the given 2D poses and anatomical constraints of humans. Experiments on the Yoga dataset demonstrate that our method achieves better cross-domain performance compared to state-of-the-art supervised and unsupervised 3D pose estimation methods, including extreme human poses where 3D supervision is unavailable. Code is available at: \url{https://github.com/asaa0001/c-MAS}.\looseness=-1
\end{abstract}

\section{Introduction}
\label{sec:intro}
3D human pose estimation is essential for numerous applications of computer vision (CV). However, the direct supervision of 3D human pose estimation networks for general purposes is still challenging due to the limited availability and domain specificity of 3D human pose datasets, such as Human3.6M~\cite{Human3.6M}.
To alleviate the difficulty of direct supervision, 3D pose estimation methods are typically divided into two sub-problems: 2D pose estimation and 2D-3D lifting. 
The 2D pose estimation stage detects 2D joint coordinates in the given image via pre-trained models on large-scale 2D datasets, such as OpenPose~\cite{OpenPose} and AlphaPose~\cite{AlphaPose}, 
followed by the lifting stage estimating the 3D joint coordinates, \ie, 3D human pose, from the 2D joint coordinates.

For 2D-3D lifting of human poses, supervised methods are widely studied (\eg,~\cite{video-to-pose3d,MotionBERT}). 
These methods directly regress 3D joint coordinates from the detected 2D joint coordinates, and are trained using existing 3D human pose datasets~\cite{Human3.6M, mpi-inf-3dhp}.
Similarly, several lifting methods are trained on multi-view 2D poses~\cite{Multi-View_Geometry} or depth sequences~\cite{Ordinal_Depth_Supervision} extracted from 3D datasets or videos.
Although lifting methods with 3D supervision achieve high accuracies for in-domain contexts, their performance notably drops for out-of-domain scenarios, including extreme poses such as those in Yoga~\cite{yoga90,li20223d}.\looseness=-1

To address this problem, unsupervised or self-supervised lifting methods, \ie, 2D-3D lifting without 3D supervision, have been studied (\eg, \cite{Geometric_Self-Supervision,2D_Projections_Alone?}). These methods have the advantage of utilizing 2D pose datasets for training, where extensive resources are readily accessible through 2D pose estimation on massive community videos that capture humans.
For example, ElePose~\cite{ElePose}, a state-of-the-art unsupervised method, estimates the 3D poses by evaluating the consistency of lifting results from multiple views. However, the consistency-based approach does not guarantee adherence to even simple constraints, such as anatomical correctness, resulting in inaccurate estimates in several cases, as shown in \fref{fig:teaser}.\looseness=-1

Extending the unsupervised methods, we propose introducing rich priors of 2D motion diffusion models (MDMs) for the 2D-3D lifting task. 
While 3D MDMs~\cite{mdm} trained on 3D pose datasets are widely used for motion generation tasks, we pre-train a \emph{2D} version of the MDM using rich 2D pose sequences in community videos. Leveraging the 2D MDM, we use the multi-view ancestral sampling (MAS)~\cite{MAS} to optimize 3D poses following the 2D diffusion prior by searching for adequate noises from the noise space. 
Unlike vanilla MAS proposed in \cite{MAS}, which aims for pure 3D generation tasks, we extend the framework to deal with the condition from the given 2D human pose, proposing a \textbf{conditional MAS (\cmas)} for the task of 3D pose estimation.
Our \cmas-based method is capable of introducing additional heuristic priors such as geometric and anatomical constraints, emphasizing the extensibility of our method.

Experiments on the Yoga90 dataset~\cite{yoga90} demonstrate that our \cmas achieves superior estimation accuracy of 3D joint coordinates to that of both state-of-the-art supervised (\ie, Video-to-Pose3D~\cite{video-to-pose3d} and MotionBERT~\cite{MotionBERT}) and unsupervised (\ie, ElePose~\cite{ElePose}) lifting methods. 
We also conduct ablation studies and analysis to clarify the effect of the additional priors, demonstrating the usefulness and effectiveness of our \cmas framework.

\vspace{1mm}
\noindent \emph{Contributions.} Our chief contributions are twofold:
\begin{itemize}
    \item We introduce a pre-trained diffusion prior of 2D MDMs for 3D human pose estimation, enabling an accurate 3D pose estimation from 2D videos without 3D supervision.\vspace{1mm}
    \item To this end, we newly propose \cmas, a multi-view sampling method of MDMs with given 2D observation and anatomical constraints as conditions. \vspace{1mm}
\end{itemize}   

\section{Related Work}
We introduce the motion diffusion prior to 3D pose estimation. We thus summarize the existing work of 3D pose estimation and diffusion-based motion synthesis.

\subsection{3D Pose Estimation from 2D Images}
3D human pose estimation has been widely studied, and recent methods can be technically classified into end-to-end methods, which directly regress the 3D joint coordinates, and 2D-3D lifting methods, which convert 2D pose sequences to 3D.\looseness=-1

\subsubsection{End-to-end methods}
Traditionally, compositional pose regression~\cite{Compositional_HumanPose_Regression} estimates 3D poses fully exploiting the dependencies among joints, rather than minimizing each joint's positional error independently. Pavlakos \etal~\cite{Coarse-To-Fine_Volumetric} propose an early attempt to formulate the 3D keypoint localization problem in a volumetric (voxel) space, thereby addressing the pose constraints inherent to many CNN-based methods and the computational cost of multiple forward passes.
Tekin~\etal~\cite{Direct_Predictio_Motion_Compensated_Sequences} introduce an approach that extracts spatiotemporal features from short image sequences and directly regresses 3D poses, leveraging temporal information from the outset instead of appending motion cues post hoc. Following a similar voxel-based representation of individual 3D poses, Reddy~\etal~\cite{tessetrack} enable multi-person 3D pose tracking by employing a 4D convolutional network that integrates temporal context. SSPnet~\cite{SSP-Net} is proposed to mitigate the trade-off between voxel resolution, memory consumption, and computation speed.\looseness=-1

End-to-end methods estimate 3D joint coordinates directly from 2D images, bypassing intermediate 2D joint detection and thereby reducing error propagation from the 2D estimation stage. However, these approaches require training data with 3D annotations, are sensitive to background and illumination variations, and tend to increase model complexity and size.\looseness=-1

\subsubsection{2D-3D lifting methods}
Lifting methods first estimate 2D joint coordinates from input images and subsequently infer 3D joint coordinates based on the obtained 2D coordinates. These approaches are divided into supervised methods that utilize 3D annotations and unsupervised methods that do not rely on the 3D ground truth.

Historically, supervised lifting methods have employed various architectures, such as fully connected networks (FCNs)~\cite{GFPose,Simple_yet_Effective}, temporal convolutional networks (TCNs)~\cite{spatio_temporal_networks_explicit_occlusion,video-to-pose3d}, and graph convolutional networks (GCNs)~\cite{Spatial-Temporal_Relationships,Optimizing_Network_Structure,Motion_Guided_3D}. In recent years, Transformer-based architectures have become predominant~\cite{MHFormer,P-STMO,MixSTE,Spatial_and_Temporal_Transformers}, achieving state-of-the-art accuracy. A notable method in this domain is MotionBERT~\cite{MotionBERT}, which introduces the DSTformer, a unified model for 3D pose estimation, action recognition, and mesh reconstruction. Although supervised lifting methods yield highly precise 3D joint estimates, they require large-scale datasets with 3D annotations for training.

\subsubsection{Unsupervised approaches}
Unsupervised approaches avoid the need for 3D annotations. Drover~\etal~\cite{2D_Projections_Alone?} leverage abundant 2D video data by applying adversarial training to 2D poses randomly reprojected from estimated 3D poses. Chen~\etal~\cite{Geometric_Self-Supervision} dispense with ground-truth 3D projections and enforce that lifted 3D poses remain invariant under changes in the 2D viewpoint. To address the scale ambiguity inherent in monocular lifting, Yu~\etal~\cite{Towards_Alleviating_the_Modeling_Ambiguity} introduce bone-length optimization constraints. More recently, ElePose~\cite{ElePose} models the distribution of camera orientations via normalizing flows to mitigate accuracy degradation caused by variations in camera elevation. 
Despite their success, these methods often require multi-view capture setups, which entail significant hardware cost and deployment complexity.

Building upon the recent development of unsupervised lifting approaches, we introduce diffusion priors in 2D motion diffusion models through the \cmas framework. Our experiments show that our method achieves better 3D pose estimation accuracy compared to the state-of-the-art unsupervised method~\cite{ElePose}, as well as the supervised method~\cite{MotionBERT} on unseen domains.

\begin{figure*}[t]
    \centering
    \includegraphics[width=1.0\linewidth]{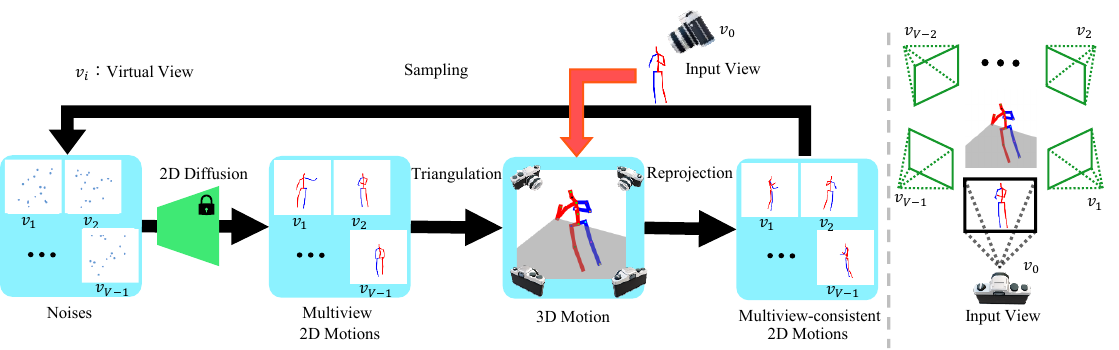}  \vspace{-7mm}
\caption{
    \textbf{(Left)} An overview of our proposed framework. We generate 2D poses for $V-1$ virtual views from noise using a diffusion model. These generated poses, along with the input 2D pose from a reference view $v_0$, are used to triangulate an initial 3D pose. The 3D pose is then reprojected, and the noise is updated to refine the estimate. This iterative process is repeated $T$ timesteps to yield the final 3D pose.
    \textbf{(Right)} A detailed illustration of the triangulation step. The 3D pose is reconstructed by triangulating the input 2D pose from the monocular camera with the multi-view virtual 2D poses.\looseness=-1
    }\vspace{-3mm}
    \label{fig:overall}
\end{figure*}

\subsection{Motion Synthesis via Diffusion Models}

Unlike traditional human motion synthesis, which relies on autoencoders~\cite{Auto-encoding} or variational autoencoders (VAEs)~\cite{Language2pose,Diverse_and_Natural,character_motion_synthesis,TEMOS,MotionCLIP}, the majority of recent methods use diffusion models~\cite{DDPM,Diffusion_Model}. These models show the superior ability to generate diverse and high-fidelity motion sequences. 
A seminal work in diffusion-based motion generation is Human MDM~\cite{mdm}, which achieves text-to-motion synthesis. 
This approach overcomes the limited diversity of VAE-based methods and the high computational cost of standard diffusion models by utilizing a lightweight Transformer architecture. This design efficiently combines the expressive power of diffusion with crucial geometric losses, resulting in diverse, physically realistic, and controllable motion.
MDMs are widely used for downstream tasks such as motion synthesis conditioned on the surrounding environment~\cite{karunratanakul2023guided,li2023object} and multi-person interaction~\cite{ghosh2024remos,liang2024intergen,tanaka2023role}.

Despite their success, a significant limitation of 3D MDMs is the heavy reliance on large-scale 3D motion capture datasets, such as HumanML3D~\cite{HumanML3D}, MPI-INF-3DHP~\cite{mpi-inf-3dhp}, and CMU MoCap\footnote{CMU Graphics Lab Motion Capture Database, \url{https://mocap.cs.cmu.edu/}, last accessed on April 13, 2026.}. Acquiring and curating these datasets is expensive and labor-intensive, which restricts the diversity and scope of the resulting motions. 
To address this, SinMDM~\cite{SinMDM} enables the learning of non-humanoid motions from a single animation. PriorMDM~\cite{PriorMDM} and GMD~\cite{karunratanakul2023guided} use fine-tuning for specific motion tasks with few or no training samples, relying on a pre-trained MDM.

More recently, MAS~\cite{MAS} introduced a novel approach to bypass the need for 3D training data. MAS leverages a pre-trained \emph{2D} MDM, which can be trained on massive in-the-wild videos, to generate 3D motion. It achieves this by generating multiple 2D views simultaneously and enforcing 3D pose consistency between them through an ancestral sampling process that incorporates triangulation and re-projection. This method successfully generates high-quality 3D motions without training on any 3D datasets.

However, the major limitation of MAS is its unconditional nature; it generates random, albeit realistic, motions from noise and lacks a mechanism for conditioning the output. Therefore, it cannot be directly applied to 2D-3D pose lifting tasks, where the output must correspond to a given 2D pose sequence. This gap highlights the need for a method that combines the data-efficiency of 2D priors with the controllability of conditional models. To address this specific challenge, we propose \cmas.

\section{Method: 2D-3D Lifting via \cmas}
We briefly recap MAS~\cite{MAS}, followed by our proposed \cmas framework for 2D-3D lifting of human poses. \fref{fig:overall} provides the overview of our method.

\subsection{Preliminary: MAS}
MAS~\cite{MAS} was originally designed for 3D motion synthesis using an unconditional version of pre-trained 2D MDMs. These 2D MDMs are trained on motion data acquired from single-view videos, in contrast to popular 3D MDMs~\cite{mdm}, which are designed for direct 3D motion generation from text prompts. 
This approach aims to address a major bottleneck in 3D motion generation, \ie, the scarcity of 3D motion capture data. 

The core mechanism of MAS is to simultaneously denoise multiple 2D motion sequences via ancestral sampling, where each sequence represents a different camera view. For these views, we consider a pinhole camera model.
Hereafter, let $V$, $L$, and $J$ be the number of (virtual) views, the sequence length, and the number of joints, respectively.

\paragraph{Noise initialization}
To enhance the multi-view consistency, MAS initializes the shared 3D noise by sampling from a Gaussian distribution as 
\begin{align}
    \bm{\epsilon}_{3D} \sim \mathcal{N}^{L \times J \times 3}(\bm{0}, \bm{I}).
\end{align}
The 3D noise is then projected onto each 2D view using the projection function $\bm{P}(\cdot)$ to obtain the 2D noise for each view $v\in\{1,\dots,V\}$ as 
\begin{align}
    \bm{\epsilon}^{1:V} = \bm{P}(\bm{\epsilon}_{3D}, v^{1:V}).
\end{align}

\paragraph{Denoising steps}
At each denoising step of the MDMs, $t=T,\dots,1$, MAS employs the diffusion model to predict the clean (\ie, $t=0$) 2D motion $\hat{\bm{x}}^{1:V}_{0} \in \mathbb{R}^{L \times J \times 2}$ for each view, guiding the multi-view consistency using a consistency block.
This block operates in two stages: triangulation and reprojection. 

In the triangulation stage, the 2D motion predictions from each view $\hat{\bm{x}}^{1:V}_{0}$ are used to reconstruct a single, unified 3D motion sequence $\bm{X} \in \mathbb{R}^{L \times J \times 3}$ that best explains all the views. 
This is achieved by minimizing the reprojection error between the projected joints of $\bm{X}$ and the predicted 2D motions. The optimization is formulated as 
\begin{equation}
    \bm{X} = \argmin_{\bm{X}'} \sum_{v=1}^{V} \|\bm{P}(\bm{X}', v) - \hat{\bm{x}}_0^v\|^2_2,
    \label{eq:geometric_loss_mas}
\end{equation}
where $\bm{P}(\cdot, v):\mathbb{R}^{L \times J \times 3} \to \mathbb{R}^{L \times J \times 2}$ denotes the perspective projection that maps a 3D motion $\bm{X}$ to its corresponding motion from a given camera view, and $\bm{X}' \in \mathbb{R}^{L \times J \times 3}$ represents the variable of 3D motion being optimized.

After the triangulation, the optimized 3D motion $\bm{X}$ is then projected back to each of the original camera views using the projection function $\bm{P}$, resulting in a set of multi-view-consistent 2D motions $\tilde{\bm{x}}^{1:V}_0$.
The reprojected 2D motions at $t=0$, $\tilde{\bm{x}}^{1:V}_0$, are then used to sample $\bm{x}^{1:V}_{t-1}$ at the next denoising step $t-1$, \ie, sampling from the diffusion posterior distribution $q(\bm{x}^{v}_{t-1} \mid \bm{x}^{v}_{t},\bm{x}_0=\tilde{\bm{x}}^{v}_{0})$.

Finally, the clean set of 2D motions obtained at $t = 0$, $\bm{x}_{0}^{1:V}$, is triangulated into the final 3D motion output $\bm{X}$.

\subsection{Conditional MAS (\cmas)}
Vanilla MAS is an unconditional generation method (\ie, it only synthesizes novel motions from a 3D noise volume), limiting its applicability in tasks requiring the reconstruction or editing of existing motions, such as 2D-3D lifting.
We, therefore, extend the MAS framework for \emph{conditional} motion generation, which we call \cmas. 
Our key idea is to adapt the consistency block within each denoising step to guide the generation process towards a specific input 2D pose sequence. Specifically, we modify the optimization problem in the triangulation stage (Eq.~(\ref{eq:geometric_loss_mas})) to incorporate two critical elements: \textbf{1) a strong fidelity term for the input view} and \textbf{2) an anatomical constraint for physical plausibility.}

\begin{figure}[tp]
\centering
\includegraphics[width=\linewidth]{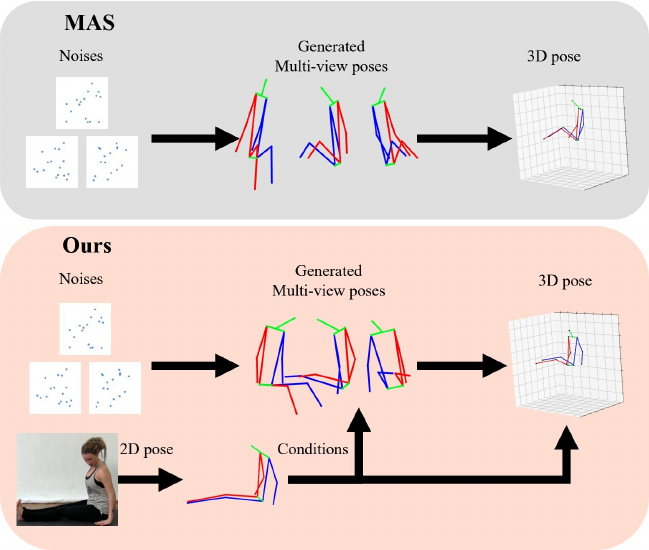}
\caption{Visualization of the multi-view 2D poses generated by the proposed method and the final triangulated 3D pose. It illustrates how consistent multi-view information, derived from the input 2D pose, enables robust 3D reconstruction.}
\label{fig:multiview_viz}
\end{figure}

\fref{fig:multiview_viz} provides an intuitive illustration for the core mechanism of our proposed approach. From a single input 2D pose, a 2D MDM generates a set of diverse yet geometrically consistent 2D poses from various virtual camera viewpoints. This rich, synthesized multi-view information provides robust cues for the triangulation step, effectively resolving the inherent depth ambiguities that challenge traditional monocular 3D lifting methods. The visualization demonstrates how leveraging a strong 2D pose prior is key to achieving a stable and accurate final 3D reconstruction.

The proposed \cmas aims to bridge the gap between unconditional generation and faithful reconstruction by leveraging an input 2D pose sequence as a powerful condition. 
Extracted from a single-view video, this sequence strictly constrains the generative process, thereby grounding the entire denoising procedure on a real-world observation.
Specifically, one of the multi-view 2D motions in the MAS framework is explicitly anchored to this input sequence, ensuring that the reconstructed 3D pose remains strictly consistent with the original viewpoint.
Driven by this anchored reference, \cmas can generate 2D poses for all the other virtual views that are both geometrically consistent with the anchor viewpoint and plausible.

Given an input 2D pose sequence $\hat{\bm{x}}^{v_0}$ from a single viewpoint $v_0$, we reformulate the optimization at each denoising step as follows:
\begin{align}
    \bm{X} = \argmin_{\bm{X}'} (\mathcal{L}_\mathrm{geometry}(\hat{\bm{x}}^{v_0}) + \lambda_\mathrm{bone} \mathcal{L}_\mathrm{bone} ).
    \label{eq:total_loss}
\end{align}
The first term $\mathcal{L}_\mathrm{geometry}(\hat{\bm{x}}^{v_0})$ represents the geometric constraint that ensures the estimated 3D pose $\bm{X}'$ is consistent with the 2D pose predictions from all $V$ views. 
The constraint is defined as:
\begin{align}
    \mathcal{L}_\mathrm{geometry} = \sum_{v=0}^{V-1} \alpha_v \|\bm{P}(\bm{X}', v) - \hat{\bm{x}}_0^v\|^2_2,
    \label{eq:geometric_loss}
\end{align}
where $\hat{\bm{x}}_0^v$ is the initial 2D pose prediction from the MDM at the current step. 
It consists of the minimization of the re-projection error between the 2D pose $\bm{P}(\bm{X}', v)$ and the generated 2D pose $\hat{\bm{x}}_0^v$ by using the MDM for each view.
Here, $\alpha_v$ is a weight for each view, designed to balance fidelity to the input 2D pose $\hat{\bm{x}}^{v_0}$ with consistency across the generated virtual views. 
The weights are specifically set as follows:
\begin{align}
    \alpha_v =
    \begin{cases}
        w_{v_0}, & \text{if } v = v_0, \\
        w=\frac{1-w_{v_0}}{V-1}, & \text{if } v \neq v_0.
    \end{cases}
    \label{eq:weight}
\end{align}
By setting a high weight for the input view ($w_{v_0} > w$), we ensure that the resulting 3D pose $\bm{X}$ is highly consistent with the provided 2D sequence from viewpoint $v_0$.

The second term, $\mathcal{L}_\mathrm{bone}$, enforces bone length constancy throughout the pose sequence.
While the MDM prior implicitly captures basic anatomical statistics, our framework allows for the flexible integration of explicit constraints to ensure physical plausibility. 
Specifically, this is achieved by minimizing the temporal variance $\sigma^2_i$ of each bone length across all frames. The anatomical constraint is formulated as follows:
\begin{equation}
\mathcal{L}_\mathrm{bone} = \frac{1}{B} \sum_{i=1}^{B} \sigma^2_i, \quad \text{where} \quad \sigma^2_i = \frac{1}{L} \sum_{l=1}^{L} \left( b_i^{(l)} - \bar{b}_i \right)^2.
\end{equation}
Here, $B$ denotes the total number of bones, and $\bar{b}_i$ represents the mean length of the $i$-th bone across $L$ frames.

The total loss for optimization is a weighted sum of the individual terms:
\begin{equation}
    \mathcal{L} = \mathcal{L}_\mathrm{geometry} + \lambda_\mathrm{bone}\mathcal{L}_\mathrm{bone},
\end{equation}
where $\lambda_\mathrm{bone}$ is the weight for the anatomical constraint $\mathcal{L}_\mathrm{bone}$.

Note that we omit explicit temporal smoothing terms, relying instead on the temporal dynamics inherently captured by the pre-trained MDM.

\section{Experiments}
\subsection{Settings}

\subsubsection{Implementation details}
We determine the weights $\alpha_v$ for the geometric loss $\mathcal{L}_\mathrm{geometry}$ as 
\begin{align}
    \alpha_v =
    \begin{cases} 
        \frac{4}{5}, & \text{if } v = v_0, \\
        \frac{1}{30}, & \text{if } v \neq v_0,
    \end{cases}
\end{align}
based on our ablation study in Table~\ref{tab:loss_weight} (Sec.~\ref{sec:geometric}), where this setting achieved the lowest error.
We use a large weight to condition the reference view $v=v_0$, where a 2D pose sequence is given.
Our 2D diffusion model is based on an MDM and is composed of a transformer encoder with 6 attention layers, 4 heads per layer, and a latent dimension of 512. 
Consistent with the original MAS setup, this backbone supports motions of variable lengths during both training and inference---a key feature inherited by our \cmas framework. 
Regarding the optimization objective in Eq.~\eqref{eq:total_loss}, we empirically set the weight for the anatomical constraint $\lambda_\mathrm{bone}$ to $0.001$. 
To solve this, we employ the Adam optimizer~\cite{kingma2014adam} with a learning rate of $0.01$, performing $1,000$ iterations at each denoising step to update the 3D pose $\bm{X}$.

For the initial training phase, we utilize the basketball dataset provided in the MAS paper~\cite{MAS}. The model is trained for 100 diffusion steps using the Adam optimizer~\cite{kingma2014adam} with a learning rate of $10^{-4}$ and a cosine noise schedule. Following this, we fine-tune the pre-trained 2D diffusion model on 2D pose sequences acquired from in-the-wild community videos containing complex poses (\eg, yoga). During fine-tuning, we apply a reduced learning rate of $10^{-6}$ and train the network for an additional $200,000$ iterations. At inference time, our method operates at approximately 2.4 FPS on a single GPU. While this is slower than real-time baselines, it is suitable for offline analysis where accuracy is prioritized.

During sampling, the virtual camera viewpoints are fixed at a distance of 7 meters from the center of the subject. All views share a constant elevation angle of $\frac{\pi}{16}$, with their azimuth angles evenly spaced across the range $[0, 2\pi]$.

\subsubsection{Training procedure}
We train our model using 2D pose sequences acquired from videos on YouTube. In total, we collected 450 videos, with lengths ranging from 4 to 20 minutes. The collection includes 200 yoga videos, 100 stretching videos, 50 physical training videos, and 100 Pilates videos. 
For all videos, we extract 2D poses using AlphaPose~\cite{AlphaPose}. Since the raw extracted poses often include joints with low confidence scores or movements that are not smooth between frames, we apply a pre-processing pipeline to ensure the reliability of the 2D poses. Specifically, we filter out joints with a confidence score below $0.3$, which is provided by the AlphaPose estimator. 
Furthermore, we segment sequences with discontinuous motion resulting from video edits or scene changes by setting a continuity threshold, $B_\mathrm{threshold}$, to $0.5$ throughout our experiment.
\begin{equation}
    \lVert \mathbf{p}_{f} - \mathbf{p}_{f-1} \rVert_2 > B_\mathrm{threshold},
\end{equation}
where $\mathbf{p}_{f}$ denotes the 2D pose at frame $f$.
Finally, to further smooth the pose sequences, we apply a space-time Gaussian filter to each joint coordinate $\mathbf{p}_{f}$. This pre-processing minimizes the dependency on the specific upstream 2D estimator.

\subsubsection{Baselines}
We compare our method with state-of-the-art baselines, including both supervised and unsupervised methods. For supervised lifting approaches, we use Video-to-Pose3D~\cite{video-to-pose3d} and MotionBERT~\cite{MotionBERT}, which are trained on the Human3.6M~\cite{Human3.6M} dataset with 3D annotations to explicitly evaluate cross-domain generalization on OOD data. For the unsupervised lifting method, we use ElePose~\cite{ElePose} representing the state-of-the-art in generative unsupervised lifting. Regarding implementations, we use a community codebase\footnote{Video-to-Pose3D, \url{https://github.com/zh-plus/video-to-pose3D}, last accessed on April 13, 2026.} for Video-to-Pose3D, the official pre-trained model for ElePose, and the codebases provided in the authors' GitHub repositories for the other baseline methods.

\subsubsection{Evaluation dataset}
To evaluate 3D pose estimation accuracy in extreme poses and OOD scenarios, we employ the Yoga90~\cite{yoga90} dataset. Unlike standard benchmarks such as Human3.6M~\cite{Human3.6M}, which primarily focus on daily activities like walking and sitting, Yoga90 specifically captures complex poses characterized by severe self-occlusion and unusual joint articulations. In this dataset, $5,526$ videos are annotated with 3D coordinates, and each annotation is accompanied by one of 90 distinct action labels. For our study, we evaluate 13 of the 32 available annotated joints. During the inference of our method, we extract 2D poses from the input videos using AlphaPose~\cite{AlphaPose}. To ensure evaluation reliability, we select a total of 384 videos for which the confidence of the 2D pose estimation is above $0.5$.

Note that there are only yoga-specific datasets to evaluate the unusual pose estimation tasks so far, and it is unrealistic to create a similar dataset for every kind of (unusual) human pose, making it difficult to quantitatively assess our method other than the yoga dataset. We also show a qualitative example for challenging poses beyond yoga in Sec.~\ref{sec:visual}.

\subsubsection{Metrics}
Following prior work, we adopt the mean per-joint position error (MPJPE) using the 3D ground-truth coordinates provided in the Yoga90 dataset. MPJPE measures the average Euclidean distance between the estimated joint positions and their ground-truth counterparts. It is a standard metric for performance comparison in 3D pose estimation, where a lower value indicates higher accuracy.

\begin{table}[bt]
\centering
\caption{\textbf{Quantitative Comparison:} MPJPE [mm] on Yoga90~\cite{yoga90}.}
\vspace{-2mm}
\label{tab:MPJPE}
\resizebox{\hsize}{!}{
    \begin{tabular}{lcc}
    \toprule
    Method & 3D Supervision & MPJPE $\downarrow$ \\
    \midrule
    MotionBERT~\cite{MotionBERT} & \checkmark & 128.930 \\
    Video-to-Pose3D~\cite{video-to-pose3d} & \checkmark & 170.837 \\
    ElePose~\cite{ElePose} & -- & 129.089\\
    \textbf{Ours} & -- & \textbf{113.370} \\
    \bottomrule
    \end{tabular}
}
\end{table}

\begin{table}[bt]
\centering
\caption{\textbf{Ablation Study} on the number of views.}\vspace{-2mm}
\resizebox{\hsize}{!}{
    \begin{tabular}{l|cccc}
    \toprule
    Views ($V$) & $V=3$ & $V=5$ & $V=7$ (Ours) & $V=9$ \\ \midrule
    MPJPE $\downarrow$ &  128.517 & 114.890 & \textbf{113.370} & 114.025 \\
    \bottomrule
    \end{tabular}
}
\end{table}

\begin{figure*}[p]
\centering
\includegraphics[width=\linewidth]{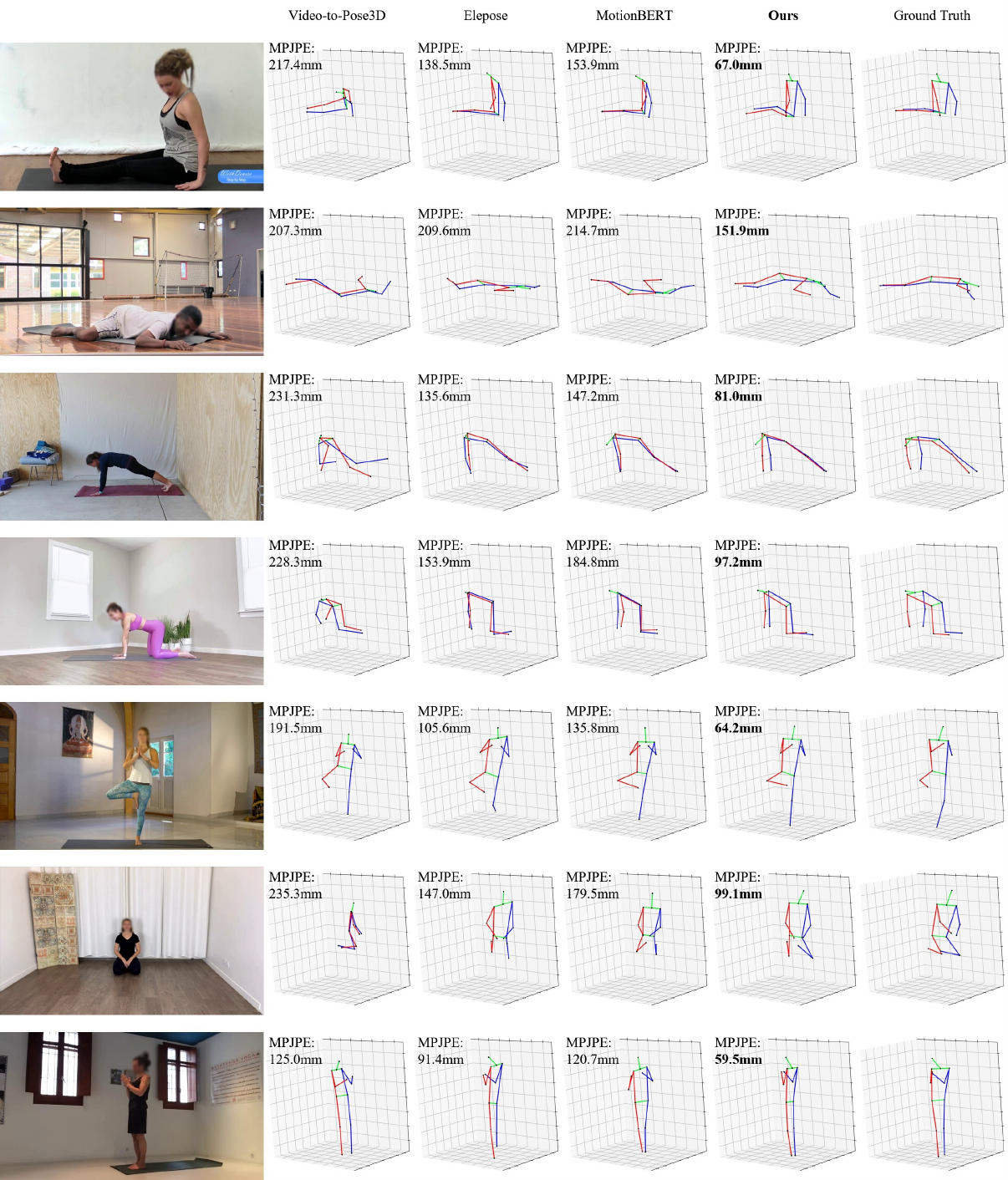}
\caption{\textbf{Qualitative comparison} of the proposed method and baseline methods on challenging sequences from the Yoga90 dataset~\cite{yoga90}. From left to right: input, Video-to-Pose3D~\cite{video-to-pose3d}, ElePose~\cite{ElePose}, MotionBERT~\cite{MotionBERT}, Ours, and the ground truth. Compared to MotionBERT and ElePose, our method consistently generates accurate and natural 3D poses, especially for poses with significant self-occlusion or unusual joint articulations.}
\label{fig:qualitative_comparison}
\end{figure*}

\subsection{Results}
\subsubsection{Quantitative results}
Table~\ref{tab:MPJPE} shows the MPJPE [mm] for each method on the Yoga90 dataset. Lower values indicate better performance.
Our method achieves the lowest MPJPE on the Yoga90 dataset, outperforming all baseline methods. It is particularly noteworthy that our approach demonstrates superior performance even against MotionBERT~\cite{MotionBERT}, a leading supervised method. 
The superiority of our method on the Yoga90 dataset, which includes complex and diverse poses, highlights its cross-domain robustness. 
Even without direct 3D supervision in the lifting stage, the results suggest the effectiveness of an approach based on 2D diffusion priors and multi-view consistency with geometric and anatomical constraints.

\subsubsection{Qualitative results}
\label{sec:qualitative_results}
\fref{fig:qualitative_comparison} provides qualitative comparisons between our method and the baselines on several challenging sequences from the Yoga90 dataset. As illustrated, our method consistently produces more plausible and accurate 3D poses, especially in cases with significant self-occlusion or unusual joint articulations. Supervised methods like MotionBERT, which are trained on standard motion capture datasets, struggle to generalize to these extreme OOD poses. Unsupervised methods like ElePose also exhibit failures, such as wrong joint articulations and self-intersection. In contrast, our method, leveraging 2D diffusion priors, generates natural and coherent poses.

\begin{figure*}[t]
\centering
\includegraphics[width=\linewidth]{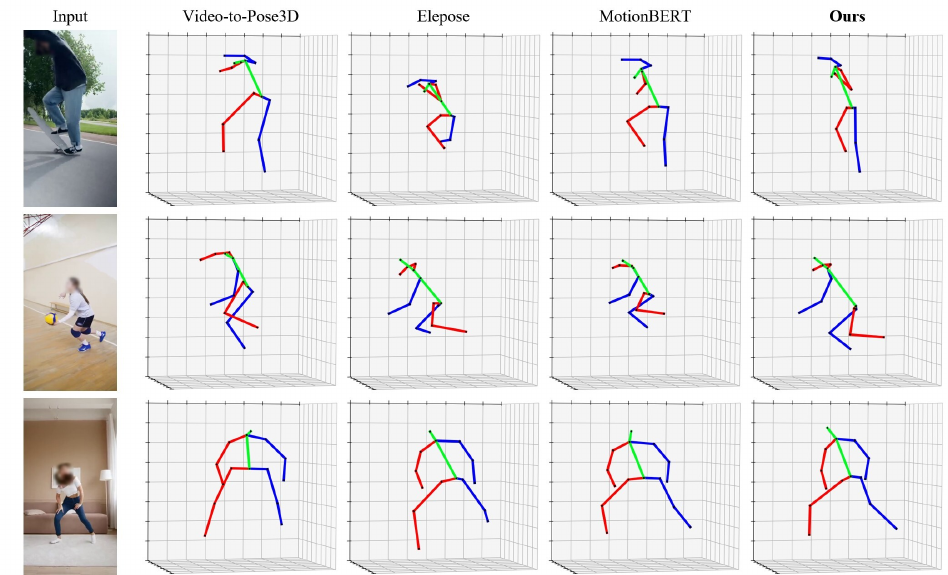}
\caption{\textbf{Visual examples of challenging poses} beyond yoga, without 3D ground truth. Compared to the baseline methods, the proposed method consistently generates accurate and natural 3D poses.} 
\label{fig:visual}
\vspace{-2mm}
\end{figure*}

\subsection{Ablation Study}
\label{sec:ablation_studies}

\begin{table}[btp]
    \centering
    \caption{\textbf{Ablation study} on the components of the proposed method on the Yoga90 dataset (MPJPE [mm]).}\vspace{-2mm}
    \label{tab:ablation_study} 
\resizebox{\hsize}{!}{
    \begin{tabular}{lc}
    \toprule
    Loss & MPJPE $\downarrow$ \\
    \midrule
    Base (Unweighted $\mathcal{L}_\mathrm{geometry}$)  & 156.760 \\
    Base + Weighted $\mathcal{L}_\mathrm{geometry}$ & 113.919 \\
    Base + Weighted $\mathcal{L}_\mathrm{geometry}$ + $\mathcal{L}_\mathrm{bone}$ (full) & \textbf{113.370} \\
    \bottomrule
    \end{tabular}
}
\end{table}

To analyze the contribution of each component to our method's performance, we conducted an ablation study. Specifically, we evaluated the impact of two main constraint terms, namely weighted geometric $\mathcal{L}_\mathrm{geometry}$ and anatomical $\mathcal{L}_\mathrm{bone}$ constraints, introduced during the triangulation optimization. The base method (without either weighted geometric $\mathcal{L}_\mathrm{geometry}$ or anatomical $\mathcal{L}_\mathrm{bone}$ constraints) uses the constant weights $\alpha_v$ for all views during the triangulation. The evaluation was performed on the Yoga90 dataset, using MPJPE~[mm] as the metric.

The results in Table~\ref{tab:ablation_study} confirm that each constraint term contributes to the performance improvement of our method. Starting from the base method with naive triangulation, introducing the weighted geometric constraint, $\mathcal{L}_\mathrm{geometry}$ with adequate weights (see \eref{eq:weight}) to enhance fidelity to the input 2D observation significantly improves the MPJPE from $156.972$~[mm] to $113.919$~[mm]. This significant gain demonstrates that prioritizing the high-confidence reference view via our weighting scheme (Eq.~\eqref{eq:weight}) effectively guides the optimization towards a more plausible 3D space.
By adding the anatomical constraint, $\mathcal{L}_\mathrm{bone}$, which imposes the physical constraint of constant bone lengths, the MPJPE is further reduced to $113.370$~[mm]. 
Since ground-truth poses are rigid, this improvement quantitatively validates the physical constraint.
The results validate the design concept of our proposed method, where it allows for easily plugging in additional constraints as loss functions.

\begin{table}[btp]
    \centering
    \caption{Effect of the weight $\alpha_{v_0}$ on the geometric loss on the Yoga90 dataset (MPJPE [mm]). We use the anatomical loss $\mathcal{L}_\mathrm{bone}$ throughout this experiment.}\vspace{-2mm}
    \label{tab:loss_weight}
\resizebox{\hsize}{!}{
    \begin{tabular}{llc}
    \toprule
    Weight $\alpha_v$ & & MPJPE $\downarrow$ \\
    \midrule
    $w_{v_0}=1/7$ & (unweighted method w/ $\mathcal{L}_\mathrm{bone}$) & 156.760\\
    $w_{v_0}=1/4$ & & 134.034 \\
    $w_{v_0}=1/3$ & & 124.646 \\
    $w_{v_0}=2/5$ & & 120.710 \\
    $w_{v_0}=1/2$ & & 117.334 \\
    $w_{v_0}=2/3$ & & 114.754 \\
    $w_{v_0}=3/4$ & & 114.112 \\
    $w_{v_0}=4/5$ & (proposed setting) &  \textbf{113.370}\\
    $w_{v_0}=9/10$ & &  114.033\\
    $w_{v_0}=1$ & (without multi-view consistency)  &  136.552\\
    \bottomrule
    \end{tabular}
}
\end{table}

\subsection{Effect of the Geometric Loss Weight}
\label{sec:geometric}
To assess the effect of the key hyperparameter in our method, we evaluate the MPJPE scores for the Yoga90 dataset by changing the weight $\alpha_{v}$, which controls the geometric loss weight for the reference view and the other (synthetic) views.
Here, we change the loss weights for the observed view $\alpha_{v_0}$, where the weights for the remaining views are computed by \eref{eq:weight}. Table~\ref{tab:loss_weight} summarizes the MPJPE [mm] by varying $\alpha_v$. Since the total number of views $V=7$, using $w_{v_0}=\frac{1}{7}$ corresponds to an unweighted baseline, while $w_{v_0}=1$ does not use any other viewpoints than the observed view (\ie, without multi-view consistency). As shown in the table, our setup with $w_{v_0}=\frac{4}{5}$ achieves the best accuracy, while constraining the observed view generally gains accuracy.\looseness=-1

\subsection{Visual Examples for Challenging Poses}
\label{sec:visual}
We show the results for challenging examples beyond the Yoga90 dataset. Since we use images from community videos or self-captured ones, the 3D ground-truth poses are inaccessible. \fref{fig:visual} shows several examples. The results consistently show that our method using 2D diffusion priors estimates reasonable 3D poses, indicating the strong generalizability of our method. Unlike ours, the methods supervised on the existing 3D motion dataset completely fail to reconstruct physically plausible 3D poses. ElePose, although being an unsupervised approach, also outputs unusual 3D poses, likely due to the lack of a 2D pose prior.

\section{Conclusions}

We introduced a novel method for estimating 3D human pose from a single-view video without 3D supervision, leveraging the rich priors from 2D motion diffusion models (MDMs) for robust 3D pose estimation. We proposed the Conditional Multi-view Ancestral Sampling (\cmas) framework, which optimizes a 3D pose by ensuring its projections from multiple virtual viewpoints adhere to the learned manifold of a pre-trained 2D MDM. A key advantage of our framework is its ability to seamlessly incorporate 2D motion input as a powerful conditioning term, as well as additional heuristic constraints such as anatomical plausibility, to guide the synthesis of consistent multi-view 2D poses. 

Although MDMs effectively capture implicit temporal relations through sequence-level processing, challenging scenarios with severe occlusions might still benefit from auxiliary temporal smoothness constraints. Incorporating such explicit priors could enhance robustness against transient 2D detection errors and produce smoother, more physically plausible motions.
Furthermore, while this paper focused on AlphaPose, adapting our method to future models is a promising direction. Our framework is inherently agnostic to the choice of the upstream 2D pose estimator.

\begin{figure}[t]
\centering
\includegraphics[width=\linewidth]{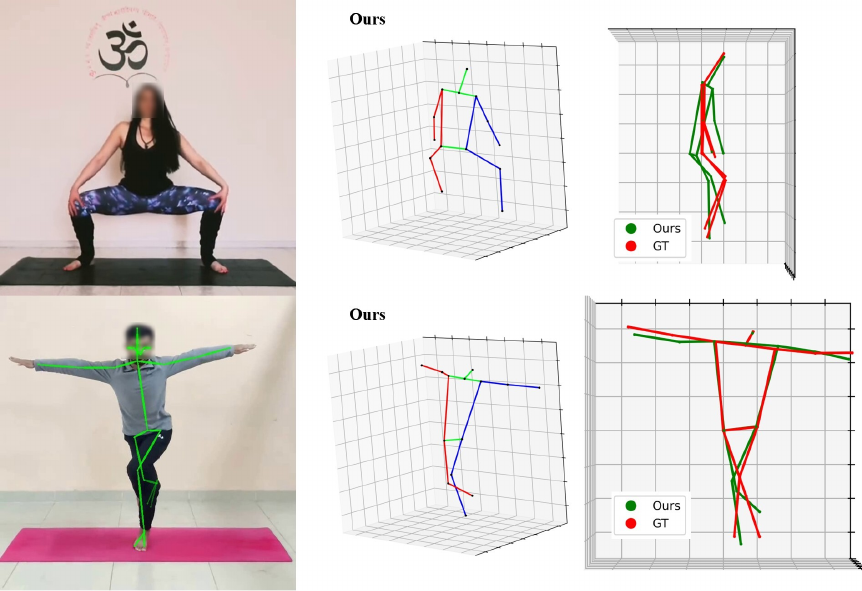}\vspace{-3mm}
\caption{Failure cases due to inherent depth ambiguity from a single view (top) and the error by the upstream 2D pose estimator (bottom).}
\vspace{-4mm}
\label{fig:limitations}
\end{figure}

\vspace{2mm}
\noindent \emph{Limitations.}
Like other 2D-to-3D lifting methods, the inherent depth ambiguity in monocular estimation can lead to incorrect reconstructions (\fref{fig:limitations}, top). We particularly found difficulties with a forward-leaning posture and with cases of extreme self-occlusion. While our anatomical constraints on bone length effectively regularize the output, they are soft constraints and may not fully correct severe initial estimation errors, leading to a geometrically consistent but anatomically imperfect solution. 
Additionally, the final 3D accuracy inherently depends on the upstream 2D pose detector, as input errors can propagate through the pipeline (\fref{fig:limitations}, bottom).

\section{Acknowledgments}
\vspace{-2mm}
This work was supported in part by the JSPS KAKENHI JP23H05491, JP25K03140, and JST FOREST JPMJFR206F.


{\small
\bibliographystyle{ieee}
\bibliography{egbib}
}

\end{document}